\documentclass[11pt,a4paper]{article}
\usepackage{times,latexsym}
\usepackage{url}
\usepackage[T1]{fontenc}
\usepackage{times}
\usepackage{latexsym}
\usepackage{booktabs}
\usepackage{multirow}
\usepackage{amsmath}
\usepackage{graphicx}
\usepackage{mdframed}
\usepackage{CJK}
\usepackage{subcaption}
\usepackage[utf8]{inputenc} 
\usepackage[T1]{fontenc}    
\usepackage{hyperref}       
\usepackage{url}            
\usepackage{booktabs}       
\usepackage{amsfonts}       
\usepackage{nicefrac}       
\usepackage{microtype}      
\usepackage{xcolor}         
\usepackage[misc,geometry]{ifsym}
\usepackage{tabularx}

\usepackage[acceptedWithA]{tacl2021v1}

\usepackage{xspace,mfirstuc,tabulary}

\newif\iftaclinstructions
\taclinstructionsfalse 
\iftaclinstructions
\renewcommand{\confidential}{}
\renewcommand{\anonsubtext}{(No author info supplied here, for consistency with
TACL-submission anonymization requirements)}
\newcommand{\instr}
\fi

\iftaclpubformat 

\else

\fi

\newcommand*{\escape}[1]{\texttt{\textbackslash#1}}

\title{GPT-4 vs. Human Translators: A Comprehensive Evaluation of Translation Quality Across Languages, Domains, and Expertise Levels}

\author{%
Jianhao Yan$^{1, 2}$\thanks{~~These authors contributed equally to this work.} \hspace{1em}
Pingchuan Yan$^{3*}$ \hspace{1em} Yulong Chen$^{4*}$ \hspace{1em} \\
\textbf{Judy Li}$^{5}$ \hspace{1em} \textbf{Xianchao Zhu}$^{5}$ \hspace{1em}\textbf{Yue Zhang}$^{2, 6, \text{\Letter}}$ \\
\centerline{\normalfont{$^1$ Zhejiang University} \quad \normalfont{$^2$ School of Engineering, Westlake University}} \\
\centerline{\normalfont{$^3$ University College London} \quad \normalfont{$^4$ University of Cambridge} \quad \normalfont{$^5$ Lan-Bridge Group}}\\
\centerline{\normalfont{$^6$ Institute of Advanced Technology, Westlake Institute for Advanced Study}} \\
\centerline{\texttt{elliottyan37@gmail.com}}
}

\date{}

\begin{document}
\maketitle
\begin{abstract}
This study comprehensively evaluates the translation quality of Large Language Models (LLMs), specifically GPT-4, against human translators of varying expertise levels across multiple language pairs and domains. Through carefully designed annotation rounds, we find that GPT-4 performs comparably to junior translators in terms of total errors made but lags behind medium and senior translators. We also observe the imbalanced performance across different languages and domains, with GPT-4's translation capability gradually weakening from resource-rich to resource-poor directions. 
In addition, we qualitatively study the translation given by GPT-4 and human translators, and find that GPT-4 translator suffers from literal translations, but human translators sometimes overthink the background information. 
To our knowledge, this study is the first to evaluate LLMs against human translators and analyze the systematic differences between their outputs, providing valuable insights into the current state of LLM-based translation and its potential limitations.
\end{abstract}

\section{Introduction}


Recent studies show that LLMs can serve as a strong translation system and a good substitute for NMT models~\cite{jiao2023chatgptgoodtranslatoryes, wang2023documentlevel, enis2024llm, huang2023towards, wu2024adapting, hendy2023good, peng2023towards}. For example, \citet{jiao2023chatgptgoodtranslatoryes} and \citet{wang2023documentlevel} find that GPT-4 can outperform commercial machine translation systems via automatic and human evaluations. Such impressive results have hastened a wide range of applications, including the use of GPT-4 for literary translation~\cite{wu2024perhaps}.

Despite their impressive capabilities, the nature of LLM output compared to human translators remains unclear. This raises two critical questions: (1) \emph{How do LLMs compare to human experts in translation quality?} and (2) \emph{Are there fundamental differences in their outputs?} These inquiries are particularly relevant in light of recent research demonstrating significant distinctions between LLM-generated and human-generated texts in general \cite{li2023deepfake, bao2023fast}. Such findings suggest that even if LLMs produce high-quality translations, their outputs may possess unique characteristics that distinguish them from human-produced translations. 

To determine where LLMs fall within the spectrum of human translation proficiency, which ranges from novice translators to seasoned professionals, we study the problem by taking the current representative LLM, i.e., GPT-4, and comparing it against human translators with different expertise.
We first conduct a preliminary study comparing human translations against GPT-4 translations, finding that \emph{even experts cannot reach a consensus on which translation is better}.
Given these findings, we take a finer-grained evaluation across different languages and domains, so that translation quality can be better calibrated and systematic differences can be measured.
Our evaluation covers three language pairs from resource-rich to resource-poor, i.e., Chinese$\leftrightarrow$English, Russian$\leftrightarrow$English, and Chinese$\leftrightarrow$Hindi, and three domains, i.e., News, Technology, and Biomedical. 
Given a source sentence, we ask junior, medium, and senior translators and GPT-4 to generate the corresponding translation in the target language. 
Then given each translation pair, we hire independent expert annotators to label the errors in the target sentence under the MQM schema~\cite{freitag2021experts}.
We find that GPT-4 reaches a comparable performance to junior translators in the perspective of total errors made, and lags behind senior ones with a considerable gap. 

Our further analyses and qualitative studies show that there are imbalanced performances for different languages and domains.
From resource-rich to resource-poor directions, GPT-4's translation capability gradually weakens. 
For resource-rich directions like Chinese$\leftrightarrow$English, GPT-4 performs comparably with junior translators and even close to medium translators, but in Chinese$\leftrightarrow$Hindi, it even lags behind our baseline system. 
The weaknesses mentioned above are also general shortcomings of large models and reflect that although large models have achieved universal translation with a focus on one language, translation between low-resource languages remains a relative weakness.

To our knowledge, we are the first to evaluate LLMs against human translators and analyze the systematic differences between LLMs and human translators.  







\section{Related Work}
\paragraph{Benchmarking LLMs}
Previous studies have benchmarked LLMs on various NLP tasks. 
\citet{xu2020clue} benchmark several LLMs on Chinese text, evaluating their Chinese ability.
\citet{ye2024benchmarking} assess LLMs through Question Answering (QA), MMLU \cite{hendrycks2021measuring}, and other metrics. From these tests, LLMs with larger scales are generally proved to be more accurate except for certain tasks. \citet{yuan2023revisiting} demonstrates that LLMs perform well in long-context understanding and are more capable with Out-of-Distribution, which means LLMs have a certain degree of generalization ability. 

Further to the MT field, \citet{jiao2023chatgpt} find that GPT-4 performed competitively with other SotA translation products. \citet{wang2023documentlevel} further investigated the capability of GPT-4 in document-level translation, the results show that GPT-4 performs better than commercial translation products and document NMT methods.
Compared to them, our work empirically shows that GPT-4 is comparable to junior human translators.

\paragraph{LLMs as Human Experts}

Due to the great capacities of GPT-4 over traditional NLP models, researchers have investigated and compared the performance of GPT-4 as human experts in multiple NLP tasks.
\citet{zhu2024benchmarking} highlight that GPT-4 and GPT-4-turbo show top performance on a Chinese financial language understanding task. 
\citet{liu2023evaluating} find the LLMs can be beneficial to biomedical NLP tasks.
\citet{goyal2022news} compare GPT models with several summarization models and humans, and find that GPT can generate summaries preferred by humans.
In AI for education area, \citet{nguyen2024using} show GPT-4's can provide teaching feedback for students. 
\citet{maloney2024comparison} find that GPT-4 shows close performance compared with human participants in coordination games.
\citet{siu2023chatgpt} show that GPT-4 is comparable to humans on technical translation tasks.
\citet{bojic2023gpt} find that GPT-4 can outperform human experts on linguistic pragmatic tasks. 
In clinical diagnostics, \citet{han2023comparative} find that GPT-4 can give comparable performance to humans, and GPT-4v (vision version) can even outperform human experts.

\paragraph{Human Evaluation for MT}
\cite{da} first propose Direct Assessment~(DA), which uses a continuous score from 0 to 100 to represent the quality of a hypothesis. 
DA has been adopted in WMT translation tasks for the past few years~\cite{wmt21,wmt22,wmt23}. 
MQM~\cite{lommel2014assessing}, the annotation used in this paper, is another widely used annotation scheme~\cite{mqm-usage-1,mqm-usage-2}. 
It requires the annotators to annotate the error span for each hypothesis and is shown to be more accurate and reliable than DA~\cite{freitag2021experts}. 
Thus, it is utilized in the metrics tasks of 2022 and 2023 WMT challenges~\cite{wmt22-metrics,wmt23-metrics}. 

\paragraph{Human Parity}
The human parity for machine translation systems is first claimed by \cite{hassan2018achieving}, which describes a comparable performance on the WMT 2017 news translation task from Chinese to English when compared to professional human translations. 
However, this claim is challenged by the following research, raising concerns about the limited scope of human parity. 
These limitations include the expertise of human evaluators~\citep{fischer-laubli-2020-whats}, the origin and quality of source sentences~\citep{toral-etal-2018-attaining,202110.0199}, the limited scenario of comparison~\citep{poibeau2022human} and difficulty of translation~\citep{graham-etal-2020-assessing}, indicating significant gaps between NMT models and the professional translators. 
In this work, we evaluate whether the SOTA LLM GPT-4 performs comparable to professional translators and what differs between human translators and LLMs. 
With the above lessons in mind, we address these limitations by hiring expert annotators, avoiding target-origin source text, manually evaluating source sentences, and covering high-resource to low-resource language pairs and various domains. 




\section{Preliminary Study}
This section presents our preliminary study. We aim to first compare GPT-4 translations with human translations qualitatively, in a coarse manner. 
Our comparison is simple and direct. We sample human-translated texts and prompt GPT-4 to translate the same source sentence. Then, we ask expert annotators to determine which translation is better. 

Particularly, to have a quick overview of the qualities of human translations against GPT-4 translations, we first utilize COMET-QE\footnote{Unbabel/wmt23-cometkiwi-da-xl} to score our in-house Chinese to English human-translated documents, and select two documents with the highest score and the lowest score. Note that our in-house translated documents are all translated by professional translators. 
In this way, we gather 40 pairs of translations from professional translators and GPT-4, respectively. 
Recent findings~\cite{freitag2021experts} have demonstrated that crowd-sourced human ratings are less reliable for high-quality MT evaluation. 
Thus, we hire six expert annotators to compare the two translations and select the better translations they find. 
We randomly shuffle the GPT-4 and human translations to prevent annotators from identifying GPT-4. 

The average win rate of GPT is 15.5/40~(36.25\%). It looks like a clear win for human translators, but when delving deeper, we find that the expert annotators have a low ratio of agreement with each other. 
In Table \ref{tab:compare_kappa}, most annotators only agree with each other at around 60\% (the baseline is 50\%) of an agreed winner at each source sentence. 
We further conduct a significance test and only annotator B finds human translation significantly better than GPT's translation and other annotators have high p-values. 
Given annotators' expertise and our task is straightforward, these results indicate that \textit{even expert annotators find it difficult to agree on which translation is better}, and GPT-generated translations might have different advantages against human-generated ones. 
These results motivate us to conduct a finer-grained and comprehensive evaluation to reveal the systematic difference between GPT-4 and human translations. 


\begin{table}[]
\centering
\resizebox{\linewidth}{!}{
\begin{tabular}{c|cccccc}
\toprule
Annotators & A     & B     & C     & D     & E     & F     
\\\midrule
A          & 100.0 & 57.5  & 65.0  & 65.0  & 62.5  & 67.5  \\
B          &    -   & 100.0 & 52.5  & 52.5  & 50.0  & 50.0  \\
C          &   -    &    -   & 100.0 & 65.0  & 82.5  & 67.5  \\
D          &   -    &    -   &   -    & 100.0 & 57.5  & 62.5  \\
E          &     -  &   -    &    -   &    -   & 100.0 & 70.0  \\
F          &  -     &    -   &      - &     -  &   -    & 100.0 \\\midrule
p-value & 1.000 & 0.038 & 0.268 & 0.081 & 0.154 & 0.875 \\\bottomrule
\end{tabular}}
\caption{Ratio(\%) of agreed winner across expert annotators and significance p-value for binomial test. P-value < 0.05 denotes a significant difference between GPT-4 and Human.}
\label{tab:compare_kappa}\end{table}

\renewcommand{\arraystretch}{1.2}
\newcommand{\centered}[1]{\begin{tabular}{l} #1 \end{tabular}}

\begin{table*}[]
\centering
\small
\begin{tabular}{|c|l|p{8cm}|}
\hline
\textbf{Type}                     & \textbf{Error Name}                 & \textit{\textbf{Explanations}}                                          \\ \hline
\multirow{6}{*}{Accuracy} & Mistranslation   & \textit{Translation does not accurately represent the source.}                    \\ \cline{2-3} 
                         & Addition                   & \textit{Information not present in the source.}                \\ \cline{2-3} 
                          & MT Hallucination & \textit{Information that has nothing related to source; or gibberish; or repeats }\\ \cline{2-3} 
                         & Omission                   & \textit{Missing content from the source. }                    \\ \cline{2-3} 
                         & Untranslated               & \textit{Not translated.}                    \\ \cline{2-3} 
                         & Wrong Name Entity and Term & \textit{Wrong usage of NE and Terminology.}                    \\ \hline
\multirow{6}{*}{Fluency} & Grammar                    & \textit{Problems with grammar of target language.}\\ \cline{2-3} 
                         & Punctuation                & \textit{Incorrect punctuation (for locale or style)}           \\ \cline{2-3} 
                         & Spelling & \textit{Incorrect spelling or capitalization.}\\ \cline{2-3} 
                          & Register & \textit{Wrong grammatical register (e.g., inappropriately informal pronouns).}    \\ \cline{2-3} 
                         & Inconsistent Style & \textit{Internal inconsistency ( not related to terminology )} \\ \cline{2-3} 
                         & Unnatural Flow             & \textit{Translations that are too literal or sound unnatural.} \\ \hline
Other                    & Non-translation            & -                                                     \\ \hline
\end{tabular}
\caption{Error category and explanations. We mainly follow the guidelines from Unbabel, and merge some errors to reduce the efforts for annotators to understand the annotation system. Concrete examples for each error category can be found in the Appendix. }
\label{tab:error_categories}
\end{table*}

\section{Main Experimental Setup}
Motivated by the results from our preliminary study, we conduct a comprehensive and fine-grained evaluation, for revealing the systematic difference between humans and GPTs. 
Specifically, we employed the widely recognized Multidimensional Quality Metrics (MQM) framework~\cite{lommel2014assessing} and compared human translators with varying levels of expertise to GPT-4.
Our evaluation spans multiple languages and domains, aiming to furnish broad insights into these comparisons.


\subsection{Data Collection}

We collect multilingual and multi-domain source sentences.
Our multilingual evaluation data contains six language directions, covering high resource to low resource, including English to Chinese, Chinese to English, English to Russian, Russian to English, English to Hindi, and Hindi to English. 

For general domain Chinese$\Leftrightarrow$English and English$\Leftrightarrow$Russian, we sample source sentences from test sets of WMT2023 and WMT2022, respectively. 
For Chinese$\Leftrightarrow$Hindi, we extract source news text from public websites.
For multi-domain evaluation data, we evaluate two domains, i.e., biomedical and technology and we evaluate Chinese to English. 
The source sentences are extracted news texts from public websites. 
We ensure that all sources are source language origin to avoid the effect of translationese.
We manually evaluate all source sentences for these tasks and ensure the source sentences are not too easy or too short. 
Finally, each task contains 200 sentences, making our evaluation a total of 1600 sentences. 


\subsection{Human Translators and Machine Translators}
We ask different human translators to translate our source sentences into the target language. 
Translators are of three different levels of expertise, categorized into junior-level, medium-level, and senior-level translators. 
The level of expertise is ranked by in-house criteria covering the translators' educational background, translation experience, and practical proficiency. See Appendix \ref{sec:expertise} for more details.
For a fair comparison, we request the experts not use machine translation or GPTs as assistance.
For all directions except Zh-Hi and Hi-Zh, we collect three human translation results from each level of expertise. For Zh-Hi and Hi-Zh, we only have medium-level and senior-level translators due to the scarcity of translators. 


Except for human translators, we use \texttt{gpt-4-1106-preview}, the current state-of-the-art large language model released by OpenAI and \texttt{Seamless M4T}~\cite{seamless2023} as the representative of traditional machine translations to complement our experiments.
We directly prompt GPT-4 to obtain the translation, as it is the most common practice for normal users, the easiest to reproduce, and to avoid confusion by various techniques.

\subsection{Prompt Search}
Previous study~\cite{zhao2021calibrate,liu2023pre} shows that different prompts with LLMs can result in distinctive performance. 
Thus, we collect three candidate prompts used in previous research~\cite{xu2023paradigm,jiao2023chatgptgoodtranslatoryes} and use COMET-QE~\cite{rei2020comet} to select the best prompt to make the best use of GPT-4, as shown in Table \ref{tab:prompt_search}.
In particular, we use these three prompts to prompt GPT-4 to translate 100 source sentences in our Chinese-to-English test set and adopt COMET-QE to evaluate the quality of translations. 
We find that the third prompt yields the best performance, and hence we adopt this prompt for all following experiments. 

\begin{table}[]
\small
\begin{tabular}{p{5cm}|p{1cm}}
\hline
\textbf{Prompt} & \textbf{COMET} \\\hline
Please translate the following sentence from Chinese into English. Your language and style should align with the language conventions of a native speaker. \escape{n}\textit{\{SOURCE\}}\escape{n}  & 0.775 \\\hline
You are an expert translator for translating Chinese to English. Your language and style should align with the language conventions of a native speaker. \escape{n}[Chinese]: \textit{\{SOURCE\}}\escape{n}[English]:& 0.755 \\\hline 
Please provide the English translation for these sentences. Your language and style should align with the language conventions of a native speaker. \escape{n}\textit{\{SOURCE\}}\escape{n} & \textbf{0.780} \\ 
\hline
\end{tabular}
\caption{Taking Chinese to English as an example, our three prompts and corresponding scores with COMET-QE. \textit{\{SOURCE\}} represents the source sentence to be translated. }
\label{tab:prompt_search}
\end{table}

\subsection{Annotation Protocol}
To evaluate the results of candidates' systems, we hire experts to annotate the errors of translations blindly. 
The annotation platform is Doccano~\cite{doccano}, and the error tags are made according to MQM standards. 
MQM requires the annotators to annotate the span of errors in each hypothesis. 
All hypotheses of the same source sentence are shown to the annotator together to help decide which is better. 
We have 13 error categories and two severities, as shown in Table \ref{tab:error_categories}.
Our categorization for errors mostly follows Unbabel's practice~\footnote{\url{https://help.unbabel.com/hc/en-us/articles/6444304419479-Annotation-Guidelines-Typology-3-0}} and we focus on most common error types. 
Each tag has subtags with two severities, i.e., Minor or Major.
A screenshot of the annotation system is given in Figure \ref{fig:screenshot}.

For each task, we first ask the two expert annotators to carefully read our manual and conduct a training round on the first 10 groups of translations. 
Then, we manually check these annotations to provide feedback and ask the two annotators to check their disagreements and revise their results. 
After two rounds of such training processes, we ask the annotators to finish the remaining sentences without knowing each other's results. 

After the first round of annotation, we conduct a second round to further refine the evaluation results. In particular, we hire another two experts for each task and show them the previous annotation results. They are asked to approve and make necessary modifications to previous round annotations. 
\begin{table}[]
\small
\centering
\resizebox{\linewidth}{!}{
\begin{tabular}{lrc}
\hline
\multicolumn{1}{|l|}{Task}  & \multicolumn{1}{l|}{Cohen Kappa(Segment)} & \multicolumn{1}{l|}{Krippendorffs(Span)} \\ \hline
\multicolumn{3}{|c|}{Reference, Re-Annotated by \cite{freitag2021experts}}\\ \hline
\multicolumn{1}{|l|}{WMT 2020 En-De}   & \multicolumn{1}{r|}{0.208}     &  \multicolumn{1}{r|}{0.456} \\ \hline
\multicolumn{1}{|l|}{WMT 2021 En-De}   & \multicolumn{1}{r|}{0.230}    &  \multicolumn{1}{r|}{0.501}    \\ \hline
\multicolumn{3}{|c|}{Ours}\\ \hline
\multicolumn{1}{|l|}{General Zh-En}    & \multicolumn{1}{r|}{0.257}    &  \multicolumn{1}{r|}{0.436}   \\ \hline
\multicolumn{1}{|l|}{General En-Zh}    & \multicolumn{1}{r|}{0.544}    &  \multicolumn{1}{r|}{0.579}   \\ \hline
\multicolumn{1}{|l|}{General En-Ru}    & \multicolumn{1}{r|}{0.461}    &  \multicolumn{1}{r|}{0.566}   \\ \hline
\multicolumn{1}{|l|}{General Ru-En}    & \multicolumn{1}{r|}{0.341}    &  \multicolumn{1}{r|}{0.875}   \\ \hline
\multicolumn{1}{|l|}{General Zh-Hi}    & \multicolumn{1}{r|}{0.256}    &  \multicolumn{1}{r|}{0.443}        \\ \hline
\multicolumn{1}{|l|}{General Hi-Zh}    & \multicolumn{1}{r|}{0.234}    &  \multicolumn{1}{r|}{0.495}        \\ \hline
\multicolumn{1}{|l|}{Technology Zh-En} & \multicolumn{1}{r|}{0.306}    &  \multicolumn{1}{r|}{0.581}   \\ \hline
\multicolumn{1}{|l|}{Biomedical Zh-En} & \multicolumn{1}{r|}{0.373}    &  \multicolumn{1}{r|}{{0.616}}   \\ \hline
\multicolumn{1}{|l|}{\textbf{Average}} & \multicolumn{1}{r|}{\textbf{0.321}}    &  \multicolumn{1}{r|}{\textbf{0.555}}   \\ \hline
\end{tabular}}
\caption{Cohen Kappa (segment-level) and Krippendorffs' Alpha (span-level) agreement of our annotations. }
\label{tab:final_iaa}
\end{table}

\subsection{Inter-Annotator Agreement}


Error annotation with MQM is challenging, and previous work demonstrates that the agreement scores between MQM annotations are relatively low~\cite{lommel2014assessing}. Reasons for this could be disagreement on precise spans and ambiguous error categorization~\cite{lommel2014assessing}. Despite the low agreement scores, MQM is more reliable than other evaluation protocols like Direct Assessment~\cite{freitag2021experts}.

To compute inter-annotator agreement for MQM, we employ segment-level Cohen's Kappa~\cite{cohen1960coefficient} and span-level Krippendorff's alpha~\citep{krippendorff1980validity}. For reference, we calculate the agreement on the annotated results of the 2020 and 2021 WMT English-to-German tasks by~\cite{freitag2021experts}. 
Our IAA results are shown in Table \ref{tab:final_iaa}. 
Thanks to our two-round annotation process, our IAA scores show a favorable agreement, indicating a good annotation quality. 



\section{Main Results}
\subsection{Overall Results}
\begin{figure}[!htb]
    \centering
    \begin{subfigure}{0.48\textwidth}
        \centering
        \includegraphics[width=\textwidth]{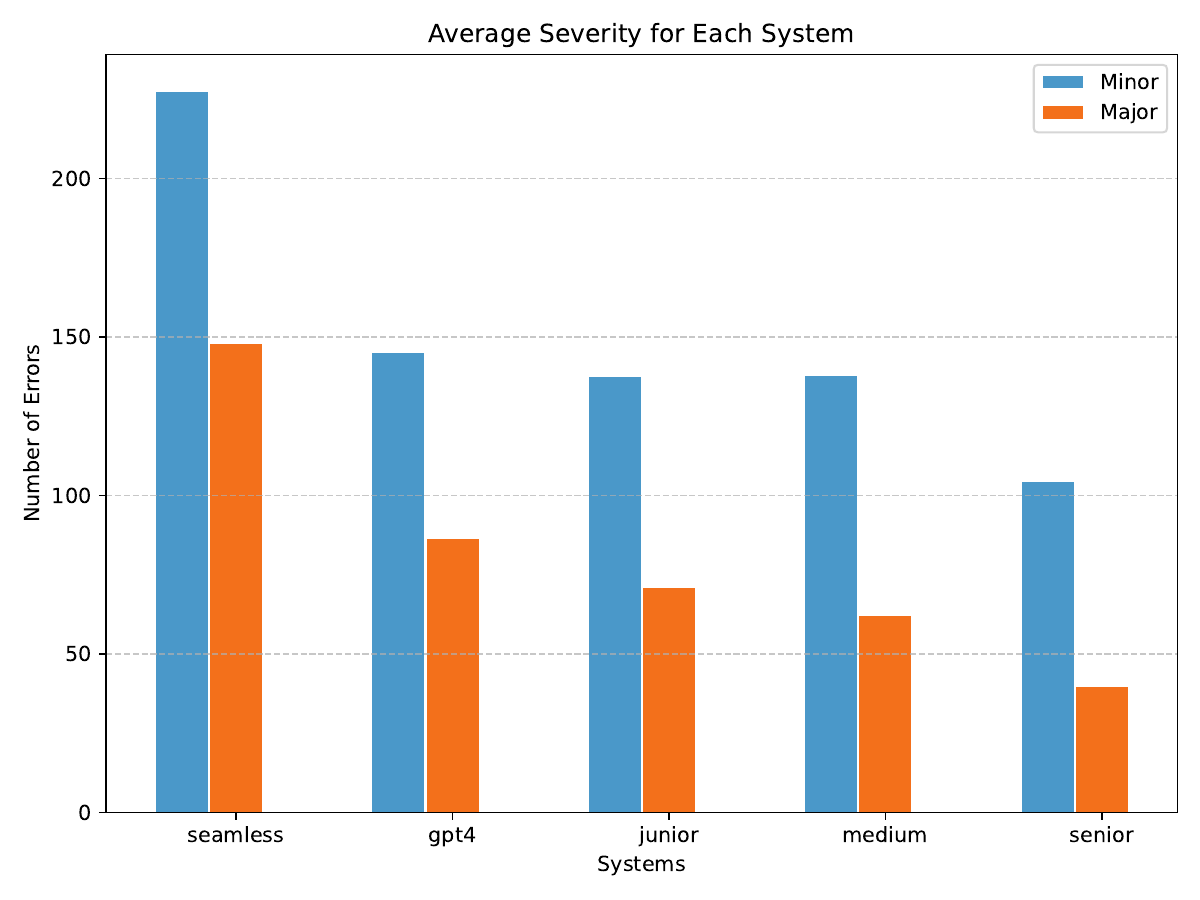}
        \label{fig:figure1}
    \end{subfigure}
    \hfill
    \begin{subfigure}{0.48\textwidth}
        \centering
        \includegraphics[width=\textwidth]{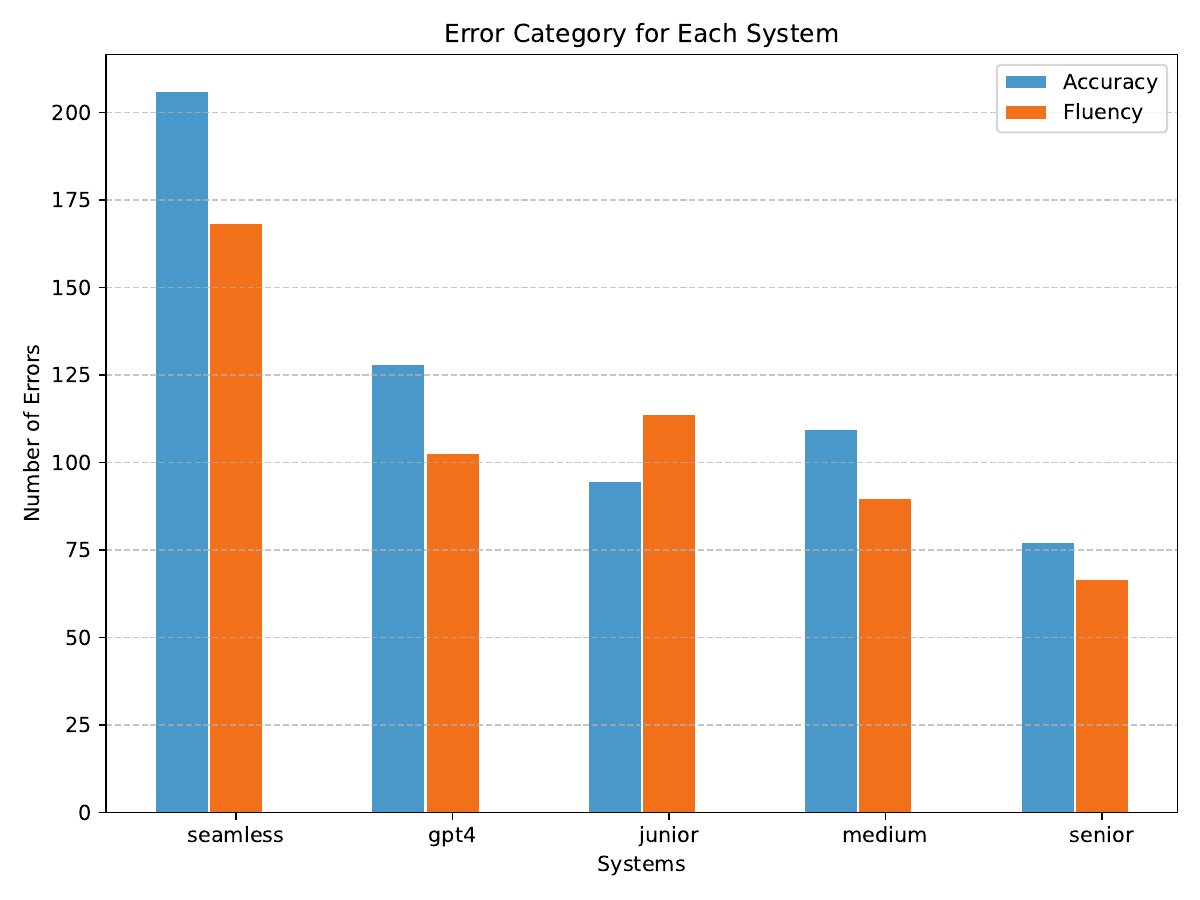}
        \label{fig:figure2}
    \end{subfigure}
    \caption{\emph{Upper}: Error severity for each system. The gray line represents the standard deviation for each system across tasks. \emph{Bottom}: Error category analysis for each system. }
    \label{fig:overall}
\end{figure}

\paragraph{Analysis of Error Severity}
The upper part of Figure \ref{fig:overall} plots the averaged number of errors of different systems and translators. 
Compared to our MT baseline~(seamless), GPT-4 has much fewer errors.
\emph{It performs almost as well as the junior-level translator at the level of total errors}, as GPT-4 is annotated with only slightly more minor and major errors than junior translators. 
However, GPT-4 still has clear performance gaps between medium or senior human translators, as it makes considerably more mistakes than experienced translators. 
To our knowledge, we are the first to report how GPT-4 is on translation against human translators. 

\paragraph{Analysis of Error Categories}
\label{sec:overall}
Furthermore, we plot the categories of errors in the bottom part of Figure \ref{fig:overall}. 
Compared with junior human translators, GPT-4 makes more errors in the accuracy of translations, which accounts for most of the disparity. Interestingly, GPT-4 surpasses junior translators in fluency issues, denoting a better capability of language usage. 

\begin{figure}[t]
    \centering
    \includegraphics[width=1.05\linewidth]{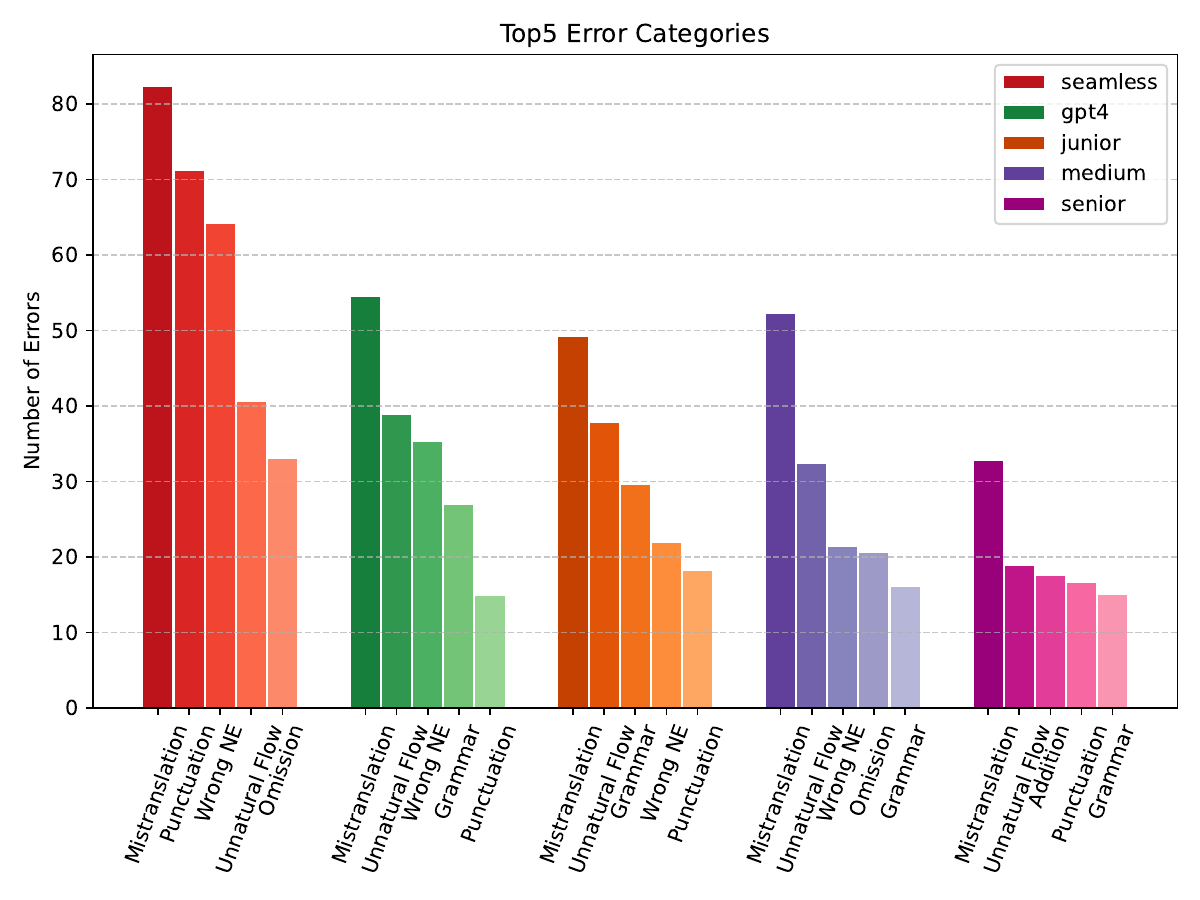}
    \caption{Top 5 categories of errors made by each system.}
    \label{fig:top5_cat}
\end{figure}

In addition, Figure \ref{fig:top5_cat} shows the top 5 categories of errors made by different systems. `Mistranslation' is the most frequent error made by all systems. Improving much over the seamless baseline, GPT-4 makes comparable numbers of `Mistranslation' with junior and medium human translators. 

For all translators, `Unnatural Flow' is among the most frequent errors. 
Seamless, GPT-4, and junior translators have similar levels of `Unnatural Flow', indicating possible issues of literal translation and not following language conventions. 
In contrast, medium and senior translators are annotated with significantly fewer errors of `Unnatural Flow'. 

In addition, we notice even though GPT-4 makes much fewer `Wrong Name Entity(NE)' errors compared to Seamless, which could be beneficial because of its huge knowledge acquired in the pre-training stage, it still has a gap compared to human translators. 

Finally, we notice that GPT-4 does not have Omission or Addition problems in its top-5 errors, whereas even senior translators have Addition errors. 


\begin{figure*}[!htb]
    \centering
    \begin{subfigure}{0.32\textwidth}
        \centering
        \includegraphics[width=\textwidth]{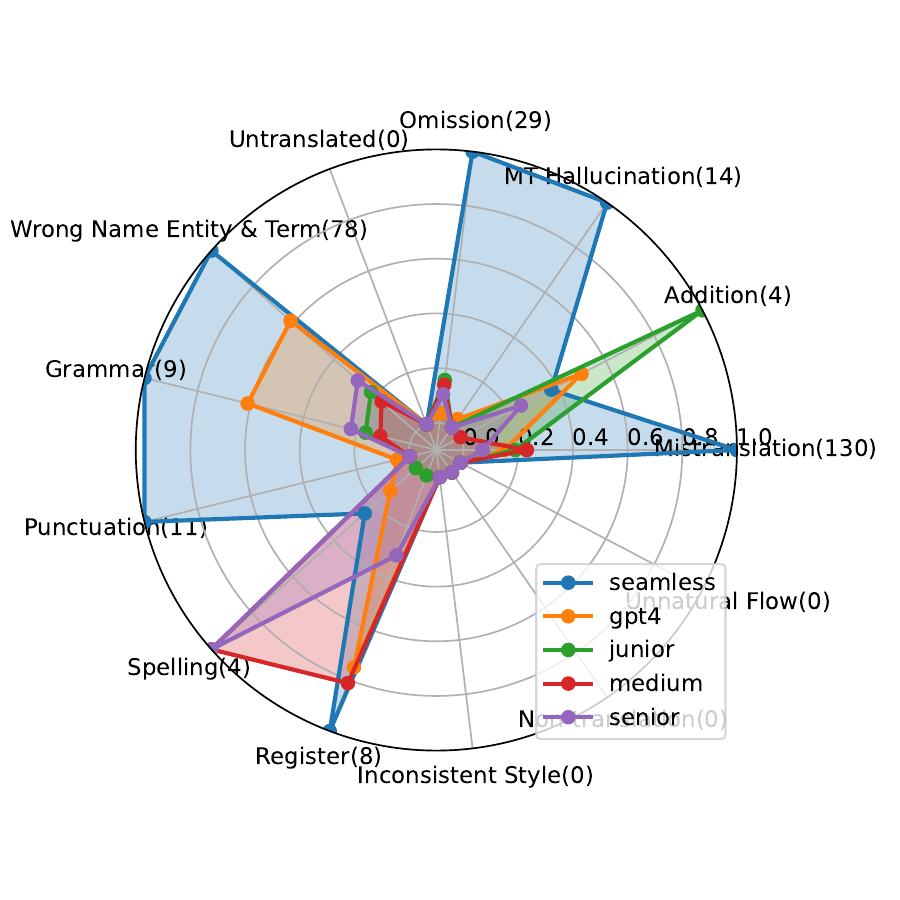}
        \caption{Chinese$\leftrightarrow$English}
        \label{fig:figure1}
    \end{subfigure}
    \hfill
    \begin{subfigure}{0.32\textwidth}
        \centering
        \includegraphics[width=\textwidth]{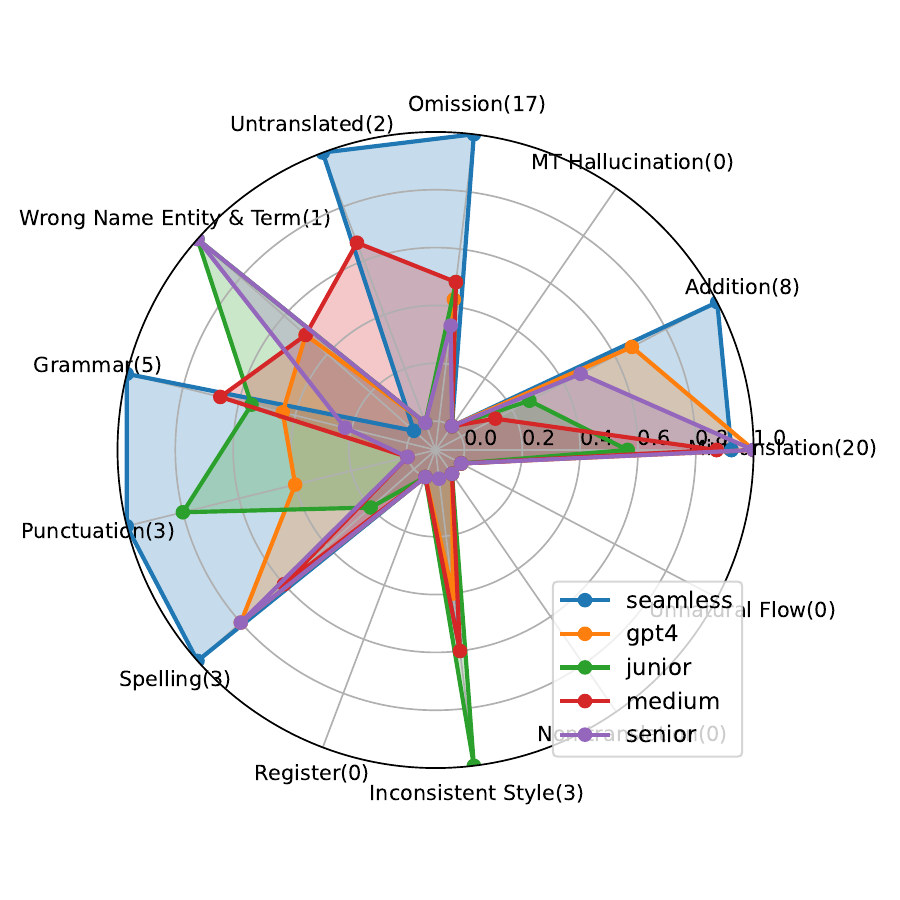}
        \caption{English$\leftrightarrow$Russian}
        \label{fig:figure1}
    \end{subfigure}
    \hfill
    \begin{subfigure}{0.32\textwidth}
        \centering
        \includegraphics[width=\textwidth]{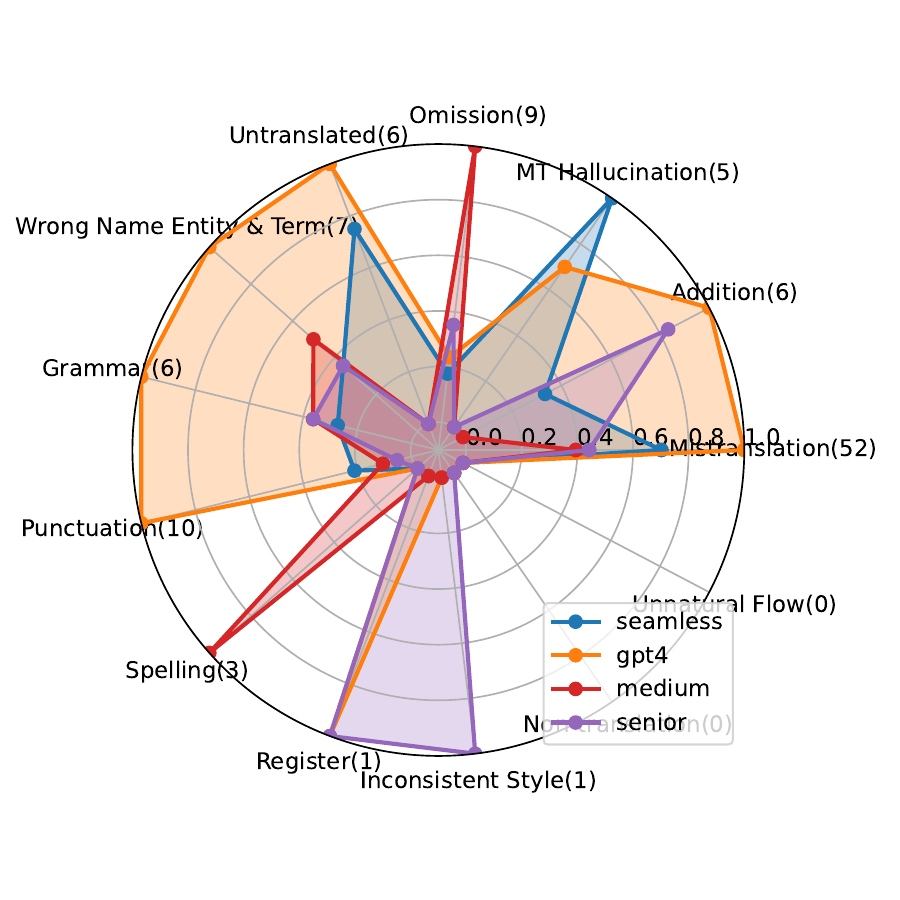}
        \caption{Chinese$\leftrightarrow$Hindi}
        \label{fig:figure2}
    \end{subfigure}
    \caption{Error category results for each language. Each sub-figure is the average over two directions. We only include `Major' errors here to highlight the most severe problems. Higher values indicate more errors and the number after each error type is the maximum number of that error. }
    \label{fig:lang}
\end{figure*}

\subsection{Detailed Results for Each Language}
In Figure \ref{fig:lang}, we present detailed results for each language pair, averaged over two directions. 
\paragraph{English-Chinese} From Figure \ref{fig:lang}(a), GPT-4 shows the great capability of translating English to Chinese and vice versa. From the radar chart, we can see that GPT-4 makes almost the same or slightly fewer semantic errors (Omission, Addition, and Mistranslation errors) than Junior and Senior translators. Especially mistranslation errors, which are generally considered most semantically detrimental, are better than junior and senior translators. For omission and addition errors, GPT-4 reaches almost the same level as senior translators. However, GPT-4 made significantly more lexical, stylistic, and grammatical errors than human translators do. The error distribution of translation of GPT-4 meets our expectations, as in the absence of reference, GPT-4 will translate unfamiliar words directly and literally instead of seeking online materials or other forms of help like human translators. Furthermore, due to the complexity and variability of Chinese, the translation of entity names or proper nouns is usually not one-to-one, two above reasons together cause the inferiority of the performance of GPT-4 in these aspects.
\paragraph{English-Russian} For the English-Russian translation tasks, GPT-4 made slightly more semantic errors but the number of mistranslation errors made by GPT-4 is almost at the same level as medium and senior translators. However, GPT-4 generally made less stylistic, grammatical, and wrong name entity \& term than junior translators. The English-Russian translation tasks are quite challenging and the performance of translators varies significantly, but GPT-4 still maintains the average level overall.
\paragraph{Hindi-Chinese}
As the low-resource language pair we evaluate, GPT-4 demonstrates the worst performance across evaluated translators. 
We observe that GPT4 is inferior to our MT baseline.
This may be due to the small portion of Hindi and Chinese corpora in its pre-training dataset.
Specifically, making the most `Mistranslation' errors of GPT-4 indicates a distance away from the language understanding of human translators. 
As a comparison, SeamlessM4T performs better in both semantic and lexical errors. 
\paragraph{Discussion}
Our results here manifest an imbalance of multilinguality for LLMs~\cite{wang2023not}. Our results imply that GPT-4 can serve as a reliable translator for resource-high such as Chinese to English but is doubtful for low-resource directions like Chinese-Hindi. In the low-resource scenario, machine translator is more reliable.

\begin{figure*}[!htb]
    \centering
    \begin{subfigure}{0.32\textwidth}
        \centering
        \includegraphics[width=\textwidth]{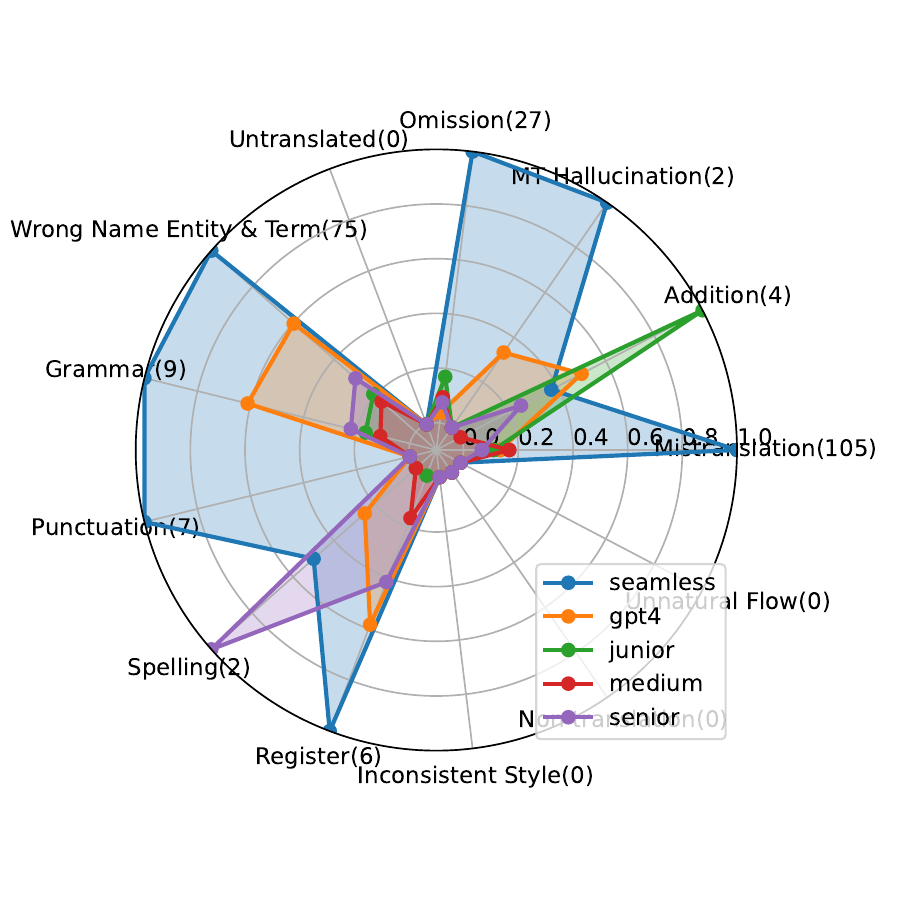}
        \caption{General news domain.}
        \label{fig:figure1}
    \end{subfigure}
    \hfill
    \begin{subfigure}{0.32\textwidth}
        \centering
        \includegraphics[width=\textwidth]{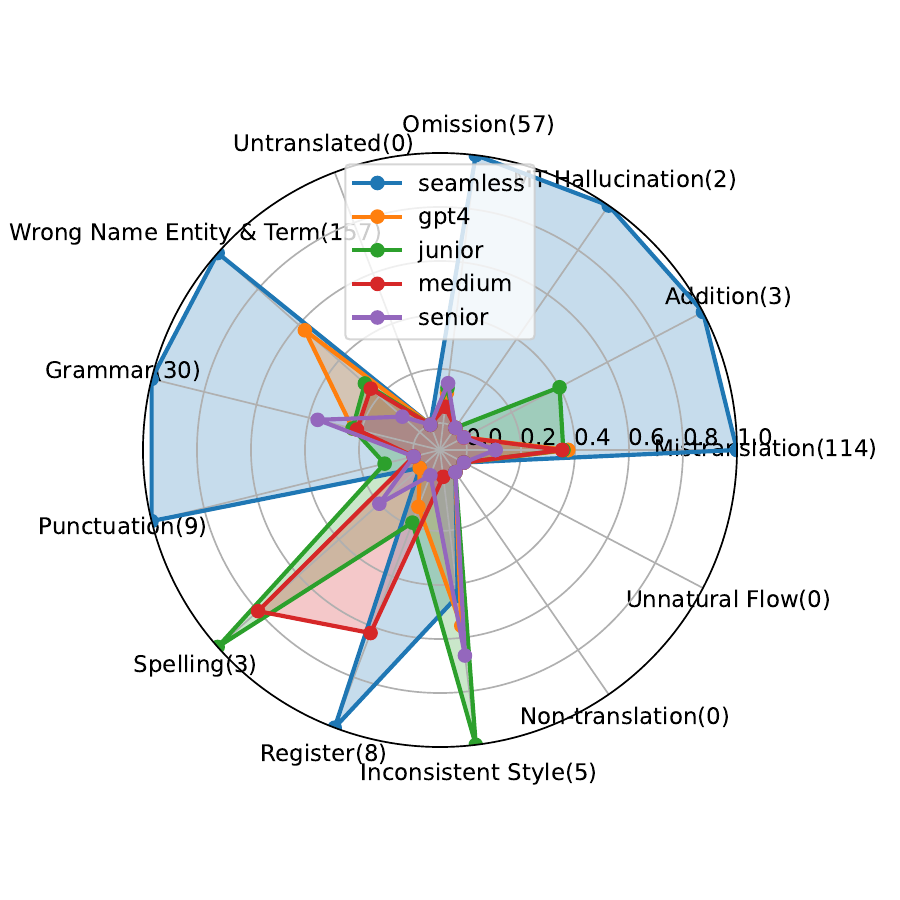}
        \caption{Technology domain.}
        \label{fig:figure1}
    \end{subfigure}
    \hfill
    \begin{subfigure}{0.32\textwidth}
        \centering
        \includegraphics[width=\textwidth]{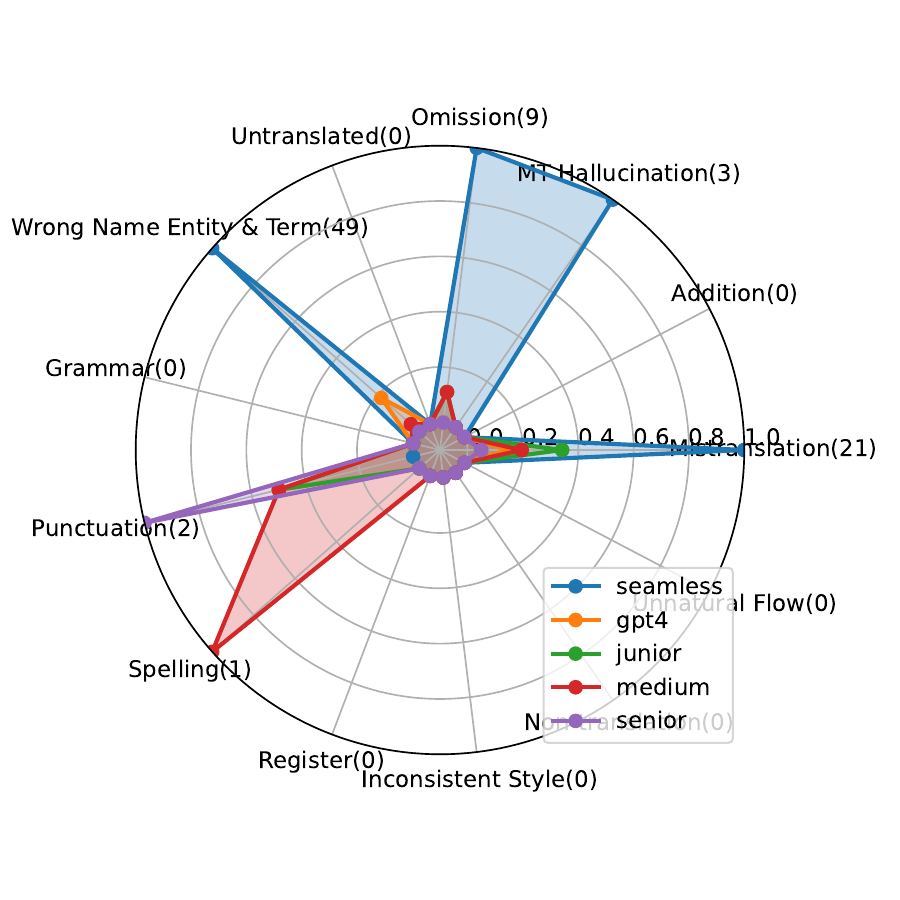}
        \caption{Biomedical domain.}
        \label{fig:figure2}
    \end{subfigure}
    \caption{Error category results for different domains in Chinese-to-English. We only include `Major' errors here to highlight the most severe problems. Higher values indicate more errors and the number in the bracket is the maximum number of that error. }
    \label{fig:domain}
\end{figure*}

\subsection{Detailed Results for Different Domains}
Figure \ref{fig:domain} presents our results for different domains in Chinese-to-English translation. 
We compare three different domains, including news, technology, and biomedical. 
\paragraph{General News Domain} GPT-4 performs worse in the news domain than human translators of three levels. The number of semantic errors made by GPT-4 is quite close to junior and medium translators. Nonetheless, GPT-4 made more lexical and grammatical errors compared to human translators. We hypothesize the reasons for the situation described above to happen are mainly because of the literariness and timeliness. Because GPT-4 is not able to access the online materials to confirm the name of a specific entity or event.
\paragraph{Technology Domain} The performance of GPT-4 is relatively close to medium-level translators. Except for the Wrong Name Entity \& Terms, GPT-4 makes almost the same or even fewer errors than medium-level translators across all aspects. Specifically, the number of semantic errors made by GPT-4 is almost the same to medium-level translators and it makes much fewer structural and grammatical errors. It means that in this field, GPT-4 might understand the original text better than junior or medium-level translators and be able to conduct a translation that is more in line with the original meaning.
\paragraph{Biomedical Domain} Similar to the technology domain, the qualities of the translations made by GPT-4 and medium-level translators stand at the same level. Despite slightly more Wrong Name Entity \& Terms errors made, GPT-4 performs better than junior and medium-level translators in other aspects.
\paragraph{Discussion}
For specific domains like technology, we show that GPT-4 is comparable with junior/medium translators. We still notice a similar imbalance issue as in the multilingual setting, but GPT-4's performance is not as sensitive as in the change of language.


\begin{table}
    \centering
    \begin{tabular}{cc}
    \toprule
    Source & \begin{CJK}{UTF8}{gbsn}巨人网络有限公司\end{CJK} \\
    GPT-4 &  Giant Network Group Inc.\\
    Human & Giant Interactive Group Inc.\\
    \bottomrule
    \end{tabular}
    \caption{Named Entity cases. }
    \label{tab:ne_case}
\end{table}

\subsection{Case Study}
We also qualitatively understand the difference between the translations given by GPT-4 and human translators. 
\paragraph{Literal Translations}
Among the error cases, the typical one is literal translations. Specifically, we find that GPT-4 sometimes translates with semantically correct, but in-native and literal translations. 
This is problematic with named entities, especially those occurring less frequently. 
As shown in Table \ref{tab:ne_case}, when not knowing the correct translation of \begin{CJK}{UTF8}{gbsn}`巨人网络有限公司'\end{CJK}, GPT-4 translates the term word by word. 
However, the issue of name entities occurs less for human translators, partially because they would google it to find the correct translation. Thus, this issue might be resolved by incorporating web-search into agent-like translation~\cite{mt-agent-1,mt-agent-2}. 

Except for named entities, we notice that the literal translation causes Unnatural Flows. 
As shown in Table \ref{tab:unnatural_case}, when translating `It's just a white screen', GPT-4 translates the phrase to \begin{CJK}{UTF8}{gbsn}`它(it)只是(is just)一个(a)白屏(white screen)'\end{CJK}, but human translator translates this phrase to `\begin{CJK}{UTF8}{gbsn}`页面显示空白(The page display is white)'\end{CJK}', which represents a preciser meaning and follows local conventions.

\begin{table}
    \centering
    \small
    \begin{tabular}{p{1.0cm}p{6cm}}
    \toprule
    Source & \underline{It's just a white screen} or it times out loading it, or the page becomes unresponsive! \\
    GPT-4 & \begin{CJK}{UTF8}{gbsn}{\color{red}它只是一个白屏}，要么是加载时超时，要么页面变得无响应了！\end{CJK} \\
    Human & \begin{CJK}{UTF8}{gbsn}{\color{green}页面要么显示空白}，要么加载超时或是无响应。\end{CJK} \\
    \bottomrule
    \end{tabular}
    \caption{Unnatural-Flow cases. Red represents the literal translation and green is more natural and native in Chinese.}
    \label{tab:unnatural_case}
\end{table}
\begin{table}
    \centering
    \small
    \begin{tabular}{p{1.0cm}p{6cm}}
    \toprule
    Source &  He has health concerns atm but \underline{we also have Daley entering his 2nd year} and is a decent safety net.\\
    GPT-4 &  \begin{CJK}{UTF8}{gbsn}他目前有健康问题，但我们还有戴利进入他的第二年，他是一个不错的安全保障。\end{CJK}\\
    Human & \begin{CJK}{UTF8}{gbsn}他目前有健康问题。不过，{\color{red}戴利两岁了}，是个不错的备选人。\end{CJK} \\
    \bottomrule
    \end{tabular}
    \caption{Human imagination cases. Red denotes the imagined part.}
    \label{tab:human_imagination}
\end{table}

\paragraph{Human Imagination}
We find human translators also have drawbacks compared to the GPT-4 translator. 
When the source sentence contains insufficient information to translate, human translators tend to fill the gap by imagination or overthinking. 
An example is given in Table \ref{tab:human_imagination}. 
The translator wrongly understands the phrase `entering his 2nd year' as Daley is a two-year-old baby, but the sentence describes a 2nd-year player for sports. 
This may be due to daily language habits, misunderstanding, or not paying attention, and could be related to the hallucination~\cite{zhang2023siren} of LLMs. 
GPT-4's literal translation helps in this, as it keeps faithful to the source sentence. This also aligns with our findings in Section \ref{sec:overall} that GPT-4 has fewer Additions or Omissions.







\section{Conclusion}
In this study, we comprehensively evaluated the translation quality of GPT-4 against human translators of varying expertise levels across multiple language pairs and domains. Our findings showed that GPT-4 performs comparably to junior translators in terms of total errors made but lags behind medium and senior translators. We also notice that GPT-4's translation capability gradually weakens from resource-rich to resource-poor language pairs. Qualitative analysis revealed that GPT-4 tends to produce more literal translations compared to human translators but suffers less from imagined information.

The results of this study demonstrate that GPT-4 has made significant strides in approaching human-level translation quality, as well as highlighting the nuanced difference between them. 
This suggests promising opportunities for collaboration and enhancement of translation workflows. As research continues to advance, we anticipate that LLMs will become increasingly valuable tools in the translation industry, working alongside human translators to improve productivity, efficiency, and overall translation quality.


\section{Limitations}
Our work is limited in the following aspects: 
(1) We benchmark GPT-4 for translation tasks, as 
it is a representative large language model and shows state-of-the-art performance for many text-based tasks. However, our evaluations can be extended to other LLMs such as Claude-3. 
(2) Our evaluation covers three languages and six directions from resource-rich to resource-poor. However, for other languages, there might be linguistic-specific phenomena that are not covered in this paper.

\bibliography{custom}

\begin{thebibliography}{51}
\expandafter\ifx\csname natexlab\endcsname\relax\def\natexlab#1{#1}\fi

\bibitem[{Bao et~al.(2023)Bao, Zhao, Teng, Yang, and Zhang}]{bao2023fast}
Guangsheng Bao, Yanbin Zhao, Zhiyang Teng, Linyi Yang, and Yue Zhang. 2023.
\newblock Fast-detectgpt: Efficient zero-shot detection of machine-generated text via conditional probability curvature.
\newblock \emph{arXiv preprint arXiv:2310.05130}.

\bibitem[{Bojic et~al.(2023)Bojic, Kovacevic, and Cabarkapa}]{bojic2023gpt}
Ljubisa Bojic, Predrag Kovacevic, and Milan Cabarkapa. 2023.
\newblock Gpt-4 surpassing human performance in linguistic pragmatics.
\newblock \emph{arXiv preprint arXiv:2312.09545}.

\bibitem[{Cohen(1960)}]{cohen1960coefficient}
Jacob Cohen. 1960.
\newblock A coefficient of agreement for nominal scales.
\newblock \emph{Educational and psychological measurement}, 20(1):37--46.

\bibitem[{Communication et~al.(2023)Communication, Barrault, Chung, Meglioli, Dale, Dong, Duquenne, Elsahar, Gong, Heffernan, Hoffman, Klaiber, Li, Licht, Maillard, Rakotoarison, Sadagopan, Wenzek, Ye, Akula, Chen, Hachem, Ellis, Gonzalez, Haaheim, Hansanti, Howes, Huang, Hwang, Inaguma, Jain, Kalbassi, Kallet, Kulikov, Lam, Li, Ma, Mavlyutov, Peloquin, Ramadan, Ramakrishnan, Sun, Tran, Tran, Tufanov, Vogeti, Wood, Yang, Yu, Andrews, Balioglu, Costa-juss\`{a}, Celebi, Elbayad, Gao, Guzm\'{a}n, Kao, Lee, Mourachko, Pino, Popuri, Ropers, Saleem, Schwenk, Tomasello, Wang, Wang, and Wang}]{seamless2023}
Seamless Communication, Lo\"{i}c Barrault, Yu-An Chung, Mariano~Cora Meglioli, David Dale, Ning Dong, Paul-Ambroise Duquenne, Hady Elsahar, Hongyu Gong, Kevin Heffernan, John Hoffman, Christopher Klaiber, Pengwei Li, Daniel Licht, Jean Maillard, Alice Rakotoarison, Kaushik~Ram Sadagopan, Guillaume Wenzek, Ethan Ye, Bapi Akula, Peng-Jen Chen, Naji~El Hachem, Brian Ellis, Gabriel~Mejia Gonzalez, Justin Haaheim, Prangthip Hansanti, Russ Howes, Bernie Huang, Min-Jae Hwang, Hirofumi Inaguma, Somya Jain, Elahe Kalbassi, Amanda Kallet, Ilia Kulikov, Janice Lam, Daniel Li, Xutai Ma, Ruslan Mavlyutov, Benjamin Peloquin, Mohamed Ramadan, Abinesh Ramakrishnan, Anna Sun, Kevin Tran, Tuan Tran, Igor Tufanov, Vish Vogeti, Carleigh Wood, Yilin Yang, Bokai Yu, Pierre Andrews, Can Balioglu, Marta~R. Costa-juss\`{a}, Onur Celebi, Maha Elbayad, Cynthia Gao, Francisco Guzm\'{a}n, Justine Kao, Ann Lee, Alexandre Mourachko, Juan Pino, Sravya Popuri, Christophe Ropers, Safiyyah Saleem, Holger Schwenk, Paden Tomasello, Changhan
  Wang, Jeff Wang, and Skyler Wang. 2023.
\newblock \href {http://arxiv.org/abs/2308.11596} {Seamlessm4t: Massively multilingual \& multimodal machine translation}.

\bibitem[{Enis and Hopkins(2024)}]{enis2024llm}
Maxim Enis and Mark Hopkins. 2024.
\newblock From llm to nmt: Advancing low-resource machine translation with claude.
\newblock \emph{arXiv preprint arXiv:2404.13813}.

\bibitem[{Farhad et~al.(2021)Farhad, Arkady, Magdalena, Ond{\v{r}}ej, Rajen, Vishrav, Costa-jussa, Cristina, Angela, Christian et~al.}]{wmt21}
Akhbardeh Farhad, Arkhangorodsky Arkady, Biesialska Magdalena, Bojar Ond{\v{r}}ej, Chatterjee Rajen, Chaudhary Vishrav, Marta~R Costa-jussa, Espa{\~n}a-Bonet Cristina, Fan Angela, Federmann Christian, et~al. 2021.
\newblock Findings of the 2021 conference on machine translation (wmt21).
\newblock In \emph{Proceedings of the Sixth Conference on Machine Translation}, pages 1--88. Association for Computational Linguistics.

\bibitem[{Feng et~al.(2024)Feng, Zhang, Li, Wu, Liao, Liu, Lang, Feng, Wu, and Liu}]{mt-agent-1}
Zhaopeng Feng, Yan Zhang, Hao Li, Bei Wu, Jiayu Liao, Wenqiang Liu, Jun Lang, Yang Feng, Jian Wu, and Zuozhu Liu. 2024.
\newblock \href {http://arxiv.org/abs/2402.16379} {Tear: Improving llm-based machine translation with systematic self-refinement}.

\bibitem[{Fischer and L{\""a}ubli(2020)}]{fischer-laubli-2020-whats}
Lukas Fischer and Samuel L{\""a}ubli. 2020.
\newblock \href {https://aclanthology.org/2020.eamt-1.23} {What's the difference between professional human and machine translation? a blind multi-language study on domain-specific {MT}}.
\newblock In \emph{Proceedings of the 22nd Annual Conference of the European Association for Machine Translation}, pages 215--224, Lisboa, Portugal. European Association for Machine Translation.

\bibitem[{Freitag et~al.(2021)Freitag, Foster, Grangier, Ratnakar, Tan, and Macherey}]{freitag2021experts}
Markus Freitag, George Foster, David Grangier, Viresh Ratnakar, Qijun Tan, and Wolfgang Macherey. 2021.
\newblock Experts, errors, and context: A large-scale study of human evaluation for machine translation.
\newblock \emph{Transactions of the Association for Computational Linguistics}, 9:1460--1474.

\bibitem[{Freitag et~al.(2023)Freitag, Mathur, Lo, Avramidis, Rei, Thompson, Kocmi, Blain, Deutsch, Stewart, Zerva, Castilho, Lavie, and Foster}]{wmt23-metrics}
Markus Freitag, Nitika Mathur, Chi-kiu Lo, Eleftherios Avramidis, Ricardo Rei, Brian Thompson, Tom Kocmi, Frederic Blain, Daniel Deutsch, Craig Stewart, Chrysoula Zerva, Sheila Castilho, Alon Lavie, and George Foster. 2023.
\newblock \href {https://doi.org/10.18653/v1/2023.wmt-1.51} {Results of {WMT}23 metrics shared task: Metrics might be guilty but references are not innocent}.
\newblock In \emph{Proceedings of the Eighth Conference on Machine Translation}, pages 578--628, Singapore. Association for Computational Linguistics.

\bibitem[{Freitag et~al.(2022)Freitag, Rei, Mathur, Lo, Stewart, Avramidis, Kocmi, Foster, Lavie, and Martins}]{wmt22-metrics}
Markus Freitag, Ricardo Rei, Nitika Mathur, Chi-kiu Lo, Craig Stewart, Eleftherios Avramidis, Tom Kocmi, George Foster, Alon Lavie, and Andr{\'e} F.~T. Martins. 2022.
\newblock \href {https://aclanthology.org/2022.wmt-1.2} {Results of {WMT}22 metrics shared task: Stop using {BLEU} {--} neural metrics are better and more robust}.
\newblock In \emph{Proceedings of the Seventh Conference on Machine Translation (WMT)}, pages 46--68, Abu Dhabi, United Arab Emirates (Hybrid). Association for Computational Linguistics.

\bibitem[{Goyal et~al.(2022)Goyal, Li, and Durrett}]{goyal2022news}
Tanya Goyal, Junyi~Jessy Li, and Greg Durrett. 2022.
\newblock News summarization and evaluation in the era of gpt-3.
\newblock \emph{arXiv preprint arXiv:2209.12356}.

\bibitem[{Graham et~al.(2013)Graham, Baldwin, Moffat, and Zobel}]{da}
Yvette Graham, Timothy Baldwin, Alistair Moffat, and Justin Zobel. 2013.
\newblock \href {https://aclanthology.org/W13-2305} {Continuous measurement scales in human evaluation of machine translation}.
\newblock In \emph{Proceedings of the 7th Linguistic Annotation Workshop and Interoperability with Discourse}, pages 33--41, Sofia, Bulgaria. Association for Computational Linguistics.

\bibitem[{Graham et~al.(2020)Graham, Federmann, Eskevich, and Haddow}]{graham-etal-2020-assessing}
Yvette Graham, Christian Federmann, Maria Eskevich, and Barry Haddow. 2020.
\newblock \href {https://doi.org/10.18653/v1/2020.findings-emnlp.375} {Assessing human-parity in machine translation on the segment level}.
\newblock In \emph{Findings of the Association for Computational Linguistics: EMNLP 2020}, pages 4199--4207, Online. Association for Computational Linguistics.

\bibitem[{Han et~al.(2023)Han, Adams, Bressem, Busch, Huck, Nebelung, and Truhn}]{han2023comparative}
Tianyu Han, Lisa~C Adams, Keno Bressem, Felix Busch, Luisa Huck, Sven Nebelung, and Daniel Truhn. 2023.
\newblock Comparative analysis of gpt-4vision, gpt-4 and open source llms in clinical diagnostic accuracy: A benchmark against human expertise.
\newblock \emph{medRxiv}, pages 2023--11.

\bibitem[{Hassan et~al.(2018)Hassan, Aue, Chen, Chowdhary, Clark, Federmann, Huang, Junczys-Dowmunt, Lewis, Li et~al.}]{hassan2018achieving}
Hany Hassan, Anthony Aue, Chang Chen, Vishal Chowdhary, Jonathan Clark, Christian Federmann, Xuedong Huang, Marcin Junczys-Dowmunt, William Lewis, Mu~Li, et~al. 2018.
\newblock Achieving human parity on automatic chinese to english news translation.
\newblock \emph{arXiv preprint arXiv:1803.05567}.

\bibitem[{Hendrycks et~al.(2021)Hendrycks, Burns, Basart, Zou, Mazeika, Song, and Steinhardt}]{hendrycks2021measuring}
Dan Hendrycks, Collin Burns, Steven Basart, Andy Zou, Mantas Mazeika, Dawn Song, and Jacob Steinhardt. 2021.
\newblock \href {http://arxiv.org/abs/2009.03300} {Measuring massive multitask language understanding}.

\bibitem[{Hendy et~al.(2023)Hendy, Abdelrehim, Sharaf, Raunak, Gabr, Matsushita, Kim, Afify, and Awadalla}]{hendy2023good}
Amr Hendy, Mohamed Abdelrehim, Amr Sharaf, Vikas Raunak, Mohamed Gabr, Hitokazu Matsushita, Young~Jin Kim, Mohamed Afify, and Hany~Hassan Awadalla. 2023.
\newblock How good are gpt models at machine translation? a comprehensive evaluation.
\newblock \emph{arXiv preprint arXiv:2302.09210}.

\bibitem[{Huang et~al.(2023)Huang, Wu, Liang, Wang, Shi, Wu, Yang, and Zhao}]{huang2023towards}
Hui Huang, Shuangzhi Wu, Xinnian Liang, Bing Wang, Yanrui Shi, Peihao Wu, Muyun Yang, and Tiejun Zhao. 2023.
\newblock Towards making the most of llm for translation quality estimation.
\newblock In \emph{CCF International Conference on Natural Language Processing and Chinese Computing}, pages 375--386. Springer.

\bibitem[{Jiao et~al.(2023{\natexlab{a}})Jiao, Wang, tse Huang, Wang, Shi, and Tu}]{jiao2023chatgptgoodtranslatoryes}
Wenxiang Jiao, Wenxuan Wang, Jen tse Huang, Xing Wang, Shuming Shi, and Zhaopeng Tu. 2023{\natexlab{a}}.
\newblock \href {http://arxiv.org/abs/2301.08745} {Is chatgpt a good translator? yes with gpt-4 as the engine}.

\bibitem[{Jiao et~al.(2023{\natexlab{b}})Jiao, Wang, tse Huang, Wang, Shi, and Tu}]{jiao2023chatgpt}
Wenxiang Jiao, Wenxuan Wang, Jen tse Huang, Xing Wang, Shuming Shi, and Zhaopeng Tu. 2023{\natexlab{b}}.
\newblock \href {http://arxiv.org/abs/2301.08745} {Is chatgpt a good translator? yes with gpt-4 as the engine}.

\bibitem[{Kim et~al.(2023)Kim, Bak, Sun, Lyu, and Lee}]{202110.0199}
Ahrii Kim, Yunju Bak, Jimin Sun, Sungwon Lyu, and Changmin Lee. 2023.
\newblock \href {https://doi.org/10.20944/preprints202110.0199.v2} {The suboptimal wmt test sets and its impact on human parity}.
\newblock \emph{Preprints}.

\bibitem[{Klubi{\v{c}}ka et~al.(2018)Klubi{\v{c}}ka, Toral, and S{\'a}nchez-Cartagena}]{mqm-usage-1}
Filip Klubi{\v{c}}ka, Antonio Toral, and V{\'\i}ctor~M S{\'a}nchez-Cartagena. 2018.
\newblock Quantitative fine-grained human evaluation of machine translation systems: a case study on english to croatian.
\newblock \emph{Machine Translation}, 32(3):195--215.

\bibitem[{Kocmi et~al.(2023)Kocmi, Avramidis, Bawden, Bojar, Dvorkovich, Federmann, Fishel, Freitag, Gowda, Grundkiewicz et~al.}]{wmt23}
Tom Kocmi, Eleftherios Avramidis, Rachel Bawden, Ond{\v{r}}ej Bojar, Anton Dvorkovich, Christian Federmann, Mark Fishel, Markus Freitag, Thamme Gowda, Roman Grundkiewicz, et~al. 2023.
\newblock Findings of the 2023 conference on machine translation (wmt23): Llms are here but not quite there yet.
\newblock In \emph{Proceedings of the Eighth Conference on Machine Translation}, pages 1--42.

\bibitem[{Kocmi et~al.(2022)Kocmi, Bawden, Bojar, Dvorkovich, Federmann, Fishel, Gowda, Graham, Grundkiewicz, Haddow et~al.}]{wmt22}
Tom Kocmi, Rachel Bawden, Ond{\v{r}}ej Bojar, Anton Dvorkovich, Christian Federmann, Mark Fishel, Thamme Gowda, Yvette Graham, Roman Grundkiewicz, Barry Haddow, et~al. 2022.
\newblock Findings of the 2022 conference on machine translation (wmt22).
\newblock In \emph{Proceedings of the Seventh Conference on Machine Translation (WMT)}, pages 1--45.

\bibitem[{Krippendorff(1980)}]{krippendorff1980validity}
Klaus Krippendorff. 1980.
\newblock Validity in content analysis.
\newblock \emph{Computerstrategien f{\"u}r die Kommunikationsanalyse}, 69:45p.

\bibitem[{Li et~al.(2023)Li, Li, Cui, Bi, Wang, Yang, Shi, and Zhang}]{li2023deepfake}
Yafu Li, Qintong Li, Leyang Cui, Wei Bi, Longyue Wang, Linyi Yang, Shuming Shi, and Yue Zhang. 2023.
\newblock Deepfake text detection in the wild.
\newblock \emph{arXiv preprint arXiv:2305.13242}.

\bibitem[{Liu et~al.(2023{\natexlab{a}})Liu, Yuan, Fu, Jiang, Hayashi, and Neubig}]{liu2023pre}
Pengfei Liu, Weizhe Yuan, Jinlan Fu, Zhengbao Jiang, Hiroaki Hayashi, and Graham Neubig. 2023{\natexlab{a}}.
\newblock Pre-train, prompt, and predict: A systematic survey of prompting methods in natural language processing.
\newblock \emph{ACM Computing Surveys}, 55(9):1--35.

\bibitem[{Liu et~al.(2023{\natexlab{b}})Liu, Zhong, Li, Zhang, Pan, Zhao, Dong, Cao, Liu, Shu et~al.}]{liu2023evaluating}
Zhengliang Liu, Tianyang Zhong, Yiwei Li, Yutong Zhang, Yi~Pan, Zihao Zhao, Peixin Dong, Chao Cao, Yuxiao Liu, Peng Shu, et~al. 2023{\natexlab{b}}.
\newblock Evaluating large language models for radiology natural language processing.
\newblock \emph{arXiv preprint arXiv:2307.13693}.

\bibitem[{Lommel et~al.(2014)Lommel, Popovic, and Burchardt}]{lommel2014assessing}
Arle Lommel, Maja Popovic, and Aljoscha Burchardt. 2014.
\newblock Assessing inter-annotator agreement for translation error annotation.
\newblock In \emph{MTE: Workshop on Automatic and Manual Metrics for Operational Translation Evaluation}, pages 31--37. Language Resources and Evaluation Conference Reykjavik.

\bibitem[{Maloney et~al.(2024)Maloney, Dal~Martello, Fei, and Ma}]{maloney2024comparison}
Laurence~T Maloney, Maria~F Dal~Martello, Vivian Fei, and Valerie Ma. 2024.
\newblock A comparison of human and gpt-4 use of probabilistic phrases in a coordination game.
\newblock \emph{Scientific reports}, 14(1):6835.

\bibitem[{Nakayama et~al.(2018)Nakayama, Kubo, Kamura, Taniguchi, and Liang}]{doccano}
Hiroki Nakayama, Takahiro Kubo, Junya Kamura, Yasufumi Taniguchi, and Xu~Liang. 2018.
\newblock \href {https://github.com/doccano/doccano} {{doccano}: Text annotation tool for human}.
\newblock Software available from https://github.com/doccano/doccano.

\bibitem[{Nguyen and Allan(2024)}]{nguyen2024using}
Ha~Nguyen and Vicki Allan. 2024.
\newblock Using gpt-4 to provide tiered, formative code feedback.
\newblock In \emph{Proceedings of the 55th ACM Technical Symposium on Computer Science Education V. 1}, pages 958--964.

\bibitem[{Peng et~al.(2023)Peng, Ding, Zhong, Shen, Liu, Zhang, Ouyang, and Tao}]{peng2023towards}
Keqin Peng, Liang Ding, Qihuang Zhong, Li~Shen, Xuebo Liu, Min Zhang, Yuanxin Ouyang, and Dacheng Tao. 2023.
\newblock Towards making the most of chatgpt for machine translation.
\newblock In \emph{Findings of the Association for Computational Linguistics: EMNLP 2023}, pages 5622--5633.

\bibitem[{Poibeau(2022)}]{poibeau2022human}
Thierry Poibeau. 2022.
\newblock On" human parity" and" super human performance" in machine translation evaluation.
\newblock In \emph{Language Resource and Evaluation Conference}.

\bibitem[{Rei et~al.(2020{\natexlab{a}})Rei, Stewart, Farinha, and Lavie}]{mqm-usage-2}
Ricardo Rei, Craig Stewart, Ana~C Farinha, and Alon Lavie. 2020{\natexlab{a}}.
\newblock \href {https://doi.org/10.18653/v1/2020.emnlp-main.213} {{COMET}: A neural framework for {MT} evaluation}.
\newblock In \emph{Proceedings of the 2020 Conference on Empirical Methods in Natural Language Processing (EMNLP)}, pages 2685--2702, Online. Association for Computational Linguistics.

\bibitem[{Rei et~al.(2020{\natexlab{b}})Rei, Stewart, Farinha, and Lavie}]{rei2020comet}
Ricardo Rei, Craig Stewart, Ana~C Farinha, and Alon Lavie. 2020{\natexlab{b}}.
\newblock Comet: A neural framework for mt evaluation.
\newblock In \emph{Proceedings of the 2020 Conference on Empirical Methods in Natural Language Processing (EMNLP)}, pages 2685--2702.

\bibitem[{Siu(2023)}]{siu2023chatgpt}
Sai~Cheong Siu. 2023.
\newblock Chatgpt and gpt-4 for professional translators: Exploring the potential of large language models in translation.
\newblock \emph{Available at SSRN 4448091}.

\bibitem[{Toral et~al.(2018)Toral, Castilho, Hu, and Way}]{toral-etal-2018-attaining}
Antonio Toral, Sheila Castilho, Ke~Hu, and Andy Way. 2018.
\newblock \href {https://doi.org/10.18653/v1/W18-6312} {Attaining the unattainable? reassessing claims of human parity in neural machine translation}.
\newblock In \emph{Proceedings of the Third Conference on Machine Translation: Research Papers}, pages 113--123, Brussels, Belgium. Association for Computational Linguistics.

\bibitem[{Wang et~al.(2023{\natexlab{a}})Wang, Lyu, Ji, Zhang, Yu, Shi, and Tu}]{wang2023documentlevel}
Longyue Wang, Chenyang Lyu, Tianbo Ji, Zhirui Zhang, Dian Yu, Shuming Shi, and Zhaopeng Tu. 2023{\natexlab{a}}.
\newblock \href {http://arxiv.org/abs/2304.02210} {Document-level machine translation with large language models}.

\bibitem[{Wang et~al.(2023{\natexlab{b}})Wang, Jiao, Huang, Dai, Huang, Tu, and Lyu}]{wang2023not}
Wenxuan Wang, Wenxiang Jiao, Jingyuan Huang, Ruyi Dai, Jen-tse Huang, Zhaopeng Tu, and Michael~R Lyu. 2023{\natexlab{b}}.
\newblock Not all countries celebrate thanksgiving: On the cultural dominance in large language models.
\newblock \emph{CoRR}.

\bibitem[{Wu et~al.(2024{\natexlab{a}})Wu, Vu, Qu, Foster, and Haffari}]{wu2024adapting}
Minghao Wu, Thuy-Trang Vu, Lizhen Qu, George Foster, and Gholamreza Haffari. 2024{\natexlab{a}}.
\newblock Adapting large language models for document-level machine translation.
\newblock \emph{arXiv preprint arXiv:2401.06468}.

\bibitem[{Wu et~al.(2024{\natexlab{b}})Wu, Yuan, Haffari, and Wang}]{wu2024perhaps}
Minghao Wu, Yulin Yuan, Gholamreza Haffari, and Longyue Wang. 2024{\natexlab{b}}.
\newblock (perhaps) beyond human translation: Harnessing multi-agent collaboration for translating ultra-long literary texts.
\newblock \emph{arXiv preprint arXiv:2405.11804}.

\bibitem[{Wu et~al.(2024{\natexlab{c}})Wu, Yuan, Haffari, and Wang}]{mt-agent-2}
Minghao Wu, Yulin Yuan, Gholamreza Haffari, and Longyue Wang. 2024{\natexlab{c}}.
\newblock \href {http://arxiv.org/abs/2405.11804} {(perhaps) beyond human translation: Harnessing multi-agent collaboration for translating ultra-long literary texts}.

\bibitem[{Xu et~al.(2023)Xu, Kim, Sharaf, and Awadalla}]{xu2023paradigm}
Haoran Xu, Young~Jin Kim, Amr Sharaf, and Hany~Hassan Awadalla. 2023.
\newblock \href {http://arxiv.org/abs/2309.11674} {A paradigm shift in machine translation: Boosting translation performance of large language models}.

\bibitem[{Xu et~al.(2020)Xu, Hu, Zhang, Li, Cao, Li, Xu, Sun, Yu, Yu et~al.}]{xu2020clue}
Liang Xu, Hai Hu, Xuanwei Zhang, Lu~Li, Chenjie Cao, Yudong Li, Yechen Xu, Kai Sun, Dian Yu, Cong Yu, et~al. 2020.
\newblock Clue: A chinese language understanding evaluation benchmark.
\newblock \emph{arXiv preprint arXiv:2004.05986}.

\bibitem[{Ye et~al.(2024)Ye, Yang, Pang, Wang, Wong, Yilmaz, Shi, and Tu}]{ye2024benchmarking}
Fanghua Ye, Mingming Yang, Jianhui Pang, Longyue Wang, Derek~F. Wong, Emine Yilmaz, Shuming Shi, and Zhaopeng Tu. 2024.
\newblock \href {http://arxiv.org/abs/2401.12794} {Benchmarking llms via uncertainty quantification}.

\bibitem[{Yuan et~al.(2023)Yuan, Chen, Cui, Gao, Zou, Cheng, Ji, Liu, and Sun}]{yuan2023revisiting}
Lifan Yuan, Yangyi Chen, Ganqu Cui, Hongcheng Gao, Fangyuan Zou, Xingyi Cheng, Heng Ji, Zhiyuan Liu, and Maosong Sun. 2023.
\newblock \href {http://arxiv.org/abs/2306.04618} {Revisiting out-of-distribution robustness in nlp: Benchmark, analysis, and llms evaluations}.

\bibitem[{Zhang et~al.(2023)Zhang, Li, Cui, Cai, Liu, Fu, Huang, Zhao, Zhang, Chen et~al.}]{zhang2023siren}
Yue Zhang, Yafu Li, Leyang Cui, Deng Cai, Lemao Liu, Tingchen Fu, Xinting Huang, Enbo Zhao, Yu~Zhang, Yulong Chen, et~al. 2023.
\newblock Siren's song in the ai ocean: a survey on hallucination in large language models.
\newblock \emph{arXiv preprint arXiv:2309.01219}.

\bibitem[{Zhao et~al.(2021)Zhao, Wallace, Feng, Klein, and Singh}]{zhao2021calibrate}
Zihao Zhao, Eric Wallace, Shi Feng, Dan Klein, and Sameer Singh. 2021.
\newblock Calibrate before use: Improving few-shot performance of language models.
\newblock In \emph{International conference on machine learning}, pages 12697--12706. PMLR.

\bibitem[{Zhu et~al.(2024)Zhu, Li, Wen, and Guo}]{zhu2024benchmarking}
Jie Zhu, Junhui Li, Yalong Wen, and Lifan Guo. 2024.
\newblock Benchmarking large language models on cflue--a chinese financial language understanding evaluation dataset.
\newblock \emph{arXiv preprint arXiv:2405.10542}.

\end{thebibliography}
\bibliographystyle{acl_natbib}

\appendix
\section{Expertise of Human Annotators}
\label{sec:expertise}
To categorize translators into junior, medium, or senior levels, we have established a comprehensive set of criteria that take into account various factors indicative of a translator's expertise and experience. These factors include the translator's educational background, particularly the prestige of the institution from which they graduated, as well as their length of service in the translation industry, the duration of their translation career, the number of translations completed, and any professional certifications they have obtained. To ensure the ongoing competence of our translators, we conduct quarterly assessments to evaluate their performance. For instance, to be classified as a senior-level translator, an individual must possess a minimum of ten years of translation experience, demonstrate exceptional proficiency by achieving a score of 99\% on our assessments, and hold the distinguished CATTI++ translation certification. By considering these stringent criteria, we aim to maintain a highly qualified and skilled pool of translators across all levels of expertise.

\section{Annotation Requirements}

\subsection{Error Types}

\begin{figure*}
    \centering
    \includegraphics[width=1.0\textwidth]{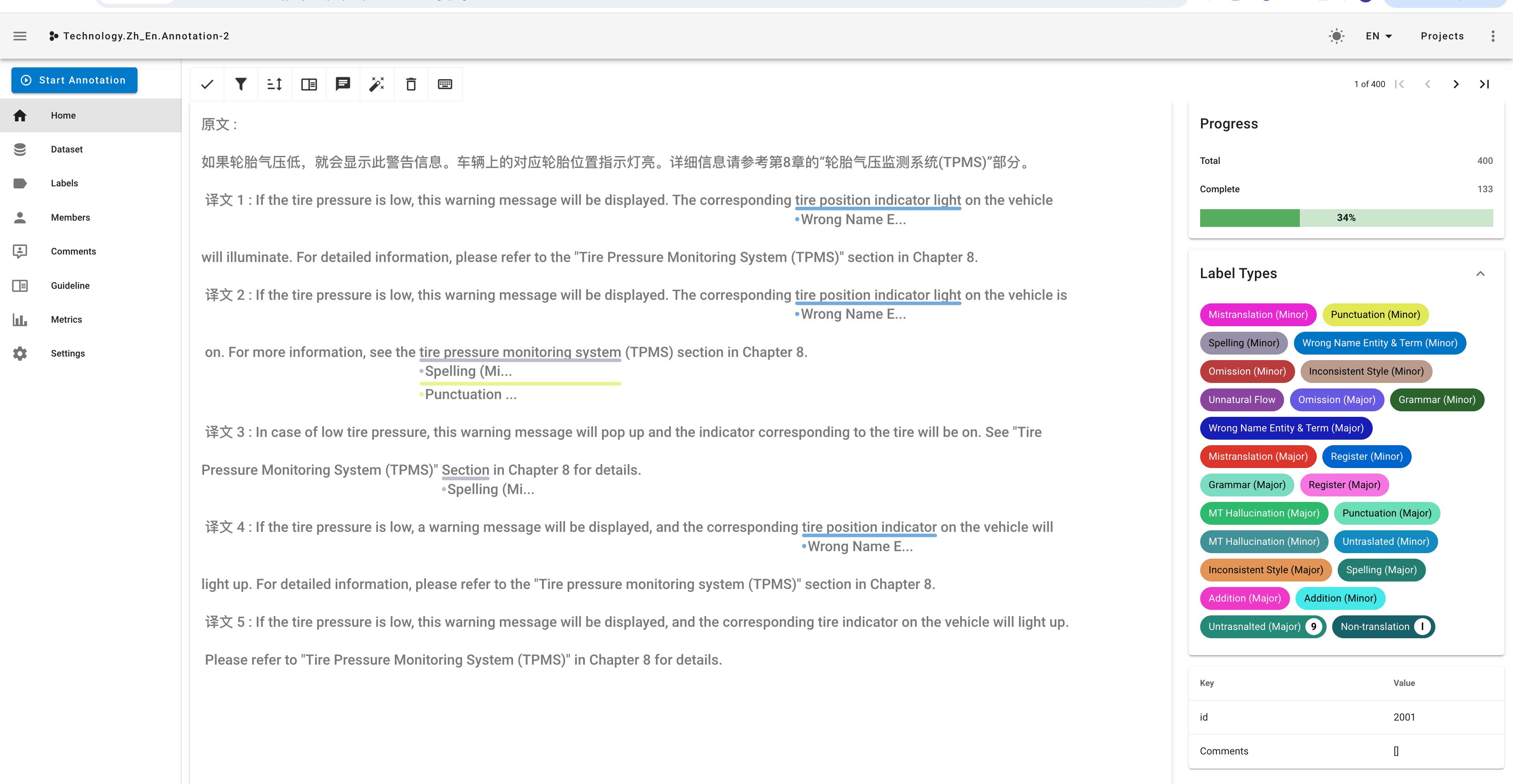}
    \caption{A screenshot of the Doccano annotation system we use. }
    \label{fig:screenshot}
\end{figure*}

Our annotation system is built upon the open-sourced doccano system~\footnote{\url{https://github.com/doccano/doccano}}. 
In Figure \ref{fig:screenshot}, we provide a screenshot of our annotation system. 
For each source sentence, outputs for different systems are given and the annotators can select spans of the text and annotate the error type and severity. 

\section{Detailed Explanation and Guidance for Each Error Types}
Our evaluation protocol largely follows the MQM criteria released by Unbabel\footnote{}. 
We provide a detailed annotation manual for annotators, including an explanation for each error type as well as illustrative examples for error types. 
It is included in the following:


\subsection{Annotation Requirements}
    The minimum unit that can be selected and annotated is a whole word, a whitespace, a punctuation mark, or an isolated character. 
    In the following example, the version in French has an extra exclamation mark, so it's necessary to annotate it as a Punctuation error:
    
    
    \emph{[EN] Thank you very much.} 
    
    \emph{[FR] Merci beaucoup!} 
    
    \begin{mdframed}
        Wrong selection $\rightarrow$Merci [beaucoup!]PUNCTUATION
        
        Correct selection $\rightarrow$Merci beaucoup[!]PUNCTUATION
    \end{mdframed} 
    
    If the issue occurs in a multiword expression, you will need to select the whole expression; if, for example, an entire sentence was translated and it shouldn't have been, you should select the entire sentence. 
    
    In the following example, we have an Unnatural Flow error: 
    
    \noindent\emph{[EN] Hi, Mary here.} 
    
    \noindent\emph{[ES] Hola, Mary aqu\'{i}.} 
    
    \begin{mdframed}
        Wrong selection $\rightarrow$ Hola, [Mary aqu\'{i}.]UNNATURAL FLOW 
        
        Correct selection $\rightarrow$ Hola, [Mary aqu\'{i}]UNNATURAL FLOW.
    \end{mdframed}

\subsection{Error Types}
\paragraph{Accuracy} 
            \begin{itemize}
                \item Mistranslation
                \begin{itemize}
                    \item Description: Translation does not accurately represent the source.
                    \item Example: 
                    \begin{mdframed}
                        \emph{[EN] It has to be done by the book.} 
                        
                        \emph{[FR] Il doit \^{e}tre fait [par le livre]MISTRANSLATION} 
                        
                        \emph{[Reason] The word-for-word translation into French doesn't work.} 
                    \end{mdframed}
                \end{itemize}
                \item Addition
                \begin{itemize}
                    \item Description: Information not present in the source.
                    \item Example: 
                    \begin{mdframed}
                        \emph{[EN] That way you can be sure that you were the one who made the changes.} 
                        
                        \emph{[ES] As\'{i} puedes estar seguro de que fuiste t\'{u} quien hizo [todos ADDItIoN los cambios.} 
                        
                        \emph{[Reason] [Todos] (meaning 'all' in Spanish) is not present in the source and it is incorrectly added in the target text.}
                    \end{mdframed}
                \end{itemize}
                \item MT Hallucination
                \begin{itemize}
                    \item Description: information that has nothing related to source; or gibberish; or repeats 
                    
                    \item Example:
                    \begin{mdframed}
                        \emph{[EN] You can send us a follow-up email at this address [EMAIL].} 
                        
                        \emph{[ES] [H\'{a}game saber si tiene alguna otra pregunta]MT HALLUCINATION.]} 
                        
                        \emph{[Reason]: The Spanish translation reads please let me know if you have any other questions and it's grammatically correct and fluent, but it has no relation at all with the source.]}
                    \end{mdframed}
                \end{itemize}
                \item Omission
                \begin{itemize}
                    \item Description: Missing content from the source.
                    \item Example:
                    \begin{mdframed}
                        \emph{[EN] We do not have much information on this.} 
                        
                        \emph{[FR] Nous ne disposons pas [] OMISSION beaucoup d'informations \`{a} ce sujet.} 
                        
                        \emph{[Reason]: The French sentence requires the preposition [de] (disposer de).}
                    \end{mdframed}
                \end{itemize}
                \item Untranslated
                \begin{itemize}
                    \item Description: Not translated.
                    \item Example:
                    \begin{mdframed}
                        \emph{[EN] How To Make Pizza Dough} 
                        
                        \emph{[FR] Comment faire de [Pizza Dough|UNTRANSLATED} 
                        
                        \emph{[Reason]: [Pizza Dough] is not a named entity and is untranslated in the French version.}
                    \end{mdframed}
                \end{itemize}
                \item Wrong Name Entity \& Term
                \begin{itemize}
                    \item Description: Wrong usage of NE and Terminology.
                    \item Example:
                    \begin{mdframed}
                        \emph{[EN] Dear Wiley, } 
                        
                        \emph{[IT] Gentile [Wilar WRONG NAMED ENTITY, } 
                        
                        \emph{[Reason]: The name in the Italian version doesn't match the original.}
                    \end{mdframed}
                \end{itemize}
            \end{itemize}
        \paragraph{Fluency} 
        \begin{itemize}
            \item Grammar
            \begin{itemize}
                \item Description: Problems with grammar of target language.
                \item Example:
                \begin{mdframed}
                    \emph{[EN] I understand that you want to check in online.} 
                    
                    \emph{[CS] ch\`{a}pu, ze se chcete [odbaven\'{i}]gRAMMaR online.} 
                    
                    \emph{[Reason]: Wrong part of speech makes the sentence ungrammatical in Czech.}
                \end{mdframed}
            \end{itemize}
            \item Punctuation
            \begin{itemize}
                \item Description: incorrect punctuation (for locale or style). 
                \item Example:
                \begin{mdframed}
                    \emph{[EN] Original copy of the Proof of Purchase or Invoice (not a screenshot): } 
                    
                    \emph{[PT] C'{o}pia original do comprovante de compra ou nota fiscal (n\~{a}o uma captura de tela)[.]PUNCTUATION} 
                    
                    \emph{[Reason]: There's a period instead of a colon in the Brazilian Portuguese version of this sentence.}
                \end{mdframed}
            \end{itemize}
            \item Spelling
            \begin{itemize}
                \item Description: incorrect spelling or capitalization.
                \item Example:
                \begin{mdframed}
                    \emph{[EN] This sort of damage is not covered under the warranty, but we will seek assistance from a higher support and see what we can do regarding this issue.} 
                    
                    \emph{[IT] Questo tipo di danno non \`{e} coperto dalla garanzia, ma chieder\`{o} comunque aiuto ai responsabili dell'assistenza per capire che cosa [Zi]SPELLING pu\`{o} fare per quanto riguarda questo problema.} 
                    
                    \emph{[Reason]: There's a typo in the sentence in Italian: the word [zi] should be [si] instead.}
                \end{mdframed}
            \end{itemize}
            \item Register
            \begin{itemize}
                \item Description: Wrong grammatical register (e.g., inappropriately informal pronouns).
                \item Example:
                \begin{mdframed}
                    \emph{[EN] Wishing you a great day ahead.} 
                    
                    \emph{[DE] Ich w\"{u}nsche [Ihnen]REGISTER einen sch\"{o}nen Tag.} 
                    
                    \emph{[Reason]: The required register for the German translation is Informal but the pronoun [Inhen] is Formal.}
                \end{mdframed}
            \end{itemize}
            \item Inconsistent Style
            \begin{itemize}
                \item Description: internal inconsistency (not related to terminology). 
                
                \item Example:
                \begin{mdframed}
                    \emph{[EN] Please click on this link. [...] This link will expire in 24 hours.} 
                    
                    \emph{[NN] Klikk p\r{a} denne [lenken].[...]Denne [linken]INCONSISTENCY utloper om 24 timer.} 
                    
                    \emph{[Reason]: Both [lenk] and [link] are correct in Norwegian, but in the same document, only one should be used. Note: this is a single error, not two}
                \end{mdframed}
            \end{itemize}
            \item Unnatural Flow
            \begin{itemize}
                \item Description: translations that are too literal or sound unnatural.
                \item Example:
                \begin{mdframed}
                    \emph{[EN] Zebras are ideal for animal matching.} 
                    
                    \emph{[DE] [Zebras sind ideal, um bestimmte Tiere zu finden]UNNATURAL FLOW.} 
                    
                    \emph{[Reason] The German translation sounds too literal, it reads like a translation, using the verb [finden] (finding) as a translation for matching. The verb matching should be translated as [detektieren] (detect) to read as if it was originally written in the target language: [Zebras sind ein ideales Beispiel zur Detektion von Wildtieren.]}
                \end{mdframed}
            \end{itemize}
        \end{itemize}
\paragraph{Other} 
        \begin{itemize}
            \item Non-translation
        \end{itemize}

\section{Extra Details}
\subsection{Translation Prompt in Preliminary Study}
In two experiments, the translation prompt we use is as follows:
\begin{itemize}
    \item Please translate the following sentences from <SRC\_LANG> to <TGT\_LANG>. Ensure line alignment across the document while maintaining the fluency of overall translation.
\end{itemize}

The prompt asks GPT4 to maintain the sentence alignment of the given document, so each sentence can be aligned back to its source sentence while being translated at the document level. 
In practice, we find most times GPT4 can follow our instructions. Occasionally, it fails to keep the sentence structure of the document and merges some sentences in one row. 
In these cases, we manually split the merged sentences. 

\subsection{Model and Decoding}
For GPT-4, we use greedy search for decoding, to ensure the reproducibility of the results. For SeamlessM4T, we use the 2.3B version of \emph{seamlessM4T\_v2\_large} and adopt beam search with beam size 5. 

\end{document}